\documentclass[letterpaper, 10 pt, conference]{ieeeconf}  

\IEEEoverridecommandlockouts                              

\usepackage{graphicx} 
\usepackage{graphics}
\usepackage{epsfig} 
\usepackage{xcolor}
\usepackage{hyperref}

\usepackage{caption}
\usepackage{subcaption}
\usepackage[utf8]{inputenc} 
\usepackage[T1]{fontenc}    
\usepackage{hyperref}       
\usepackage{url}            
\usepackage{booktabs}       

\usepackage{mathptmx} 
\usepackage{times} 
\usepackage{amsmath} 
\usepackage{amssymb}  
\usepackage{algorithm}
\usepackage{algpseudocode}
\usepackage{tabularx}
\usepackage{tabulary}
\usepackage{enumerate}
\usepackage{mathrsfs}
\usepackage{bbm}
\usepackage{hyphenat}
\usepackage{ulem}
\usepackage{multirow,hhline,color}
\usepackage{dsfont}

\usepackage[sort]{cite}
\hypersetup{breaklinks=true,colorlinks=true,linkcolor=blue,citecolor=blue}
\usepackage{tcolorbox}
\usepackage[noabbrev,capitalise,nameinlink]{cleveref}

\usepackage{color}

\newcommand{\vect}[1]{\boldsymbol{\mathbf{#1}}}

\newcommand{\ind}{\mathds{1}}
\DeclareOldFontCommand{\bf}{\normalfont\bfseries}{\mathbf}
\DeclareMathOperator*{\argmin}{argmin}
\DeclareMathOperator*{\argmax}{argmax}
\newtheorem{rem}{Remark}

\title{\LARGE \bf
A Path Planning Framework for a Flying Robot in Close Proximity of Humans\thanks{Research supported by NSF NRI initiative \#1528036.}}

\author{Hyung-Jin Yoon, Christopher Widdowson, Thiago Marinho, Ranxiao Frances Wang and Naira Hovakimyan
\thanks{Hyung-Jin Yoon, Thiago Marinho and Naira Hovakimyan are with the Department of Mechanical Science and Engineering, University of Illinois at Urbana-Champaign (UIUC), Urbana, IL 61801, USA. Christopher Widdowson and Ranxiao Wang are with Psychology Department in UIUC.
        {\tt\small \{hyoon33, widdwsn2, marinho, wang18, nhovakim\}@illinois.edu}}
}

\begin{document}

\maketitle
\thispagestyle{empty}
\pagestyle{empty}

\begin{abstract}
We present a path planning framework that takes into account the human’s safety perception in the presence of a flying robot. The framework addresses two objectives: (i) estimation of the uncertain parameters of the proposed safety perception model based on test data collected using Virtual Reality (VR) testbed, and (ii) offline optimal control computation using the estimated safety perception model. Due to the unknown factors in the human tests data, it is not suitable to use standard regression techniques that minimize the mean squared error (MSE). We propose to use a Hidden Markov model (HMM) approach where human's attention is considered as a hidden state to infer whether the data samples are relevant to learn the safety perception model. The HMM approach improved log-likelihood over the standard least squares solution. For path planning, we use Bernstein polynomials for discretization, as the resulting path remains within the convex hull of the control points, providing guarantees for deconfliction with obstacles at low computational cost. An example of optimal trajectory generation using the learned human model is presented. The optimal trajectory generated using the proposed model results in reasonable safety distance from the human. In contrast, the paths generated using the standard regression model have undesirable shapes due to overfitting. The example demonstrates that the HMM approach has robustness to the unknown factors compared to the standard MSE model.
\end{abstract}

\section{INTRODUCTION}
In the last decade, multi-rotor copters have seen immense growth in popularity, not only as a research platform, but also as a commercial and industrial device. By 2020, the market for these devices is expected to attain a value of \$11.2 billion with an annual growth of over 30\%~\cite{gartner}. The mechanical simplicity, the ability to hover and the maneuverability of these flying robots justify their use in civilian applications such as media production, inspection, and precision agriculture. The inclusion of these micro unmanned aerial vehicles (UAVs) in our day-to-day lives brings immediate benefits to society. As an example, by using fast and cheap UAVs, delivery from major retailers like Amazon and Walmart, can keep a reduced inventory resulting in cost-effective warehouse management. Additionally, by exploring their small and lightweight form factor and substantially leveraging their agility and reliability, applications in elderly care, medicine, transportation and mobile surveillance are being developed. In all these examples, it is important to fly safely near people and navigate in densely populated areas. Unlike current mobile robots that autonomously operate without considering humans, these flying collocated and cooperative robots (co-robots) are intended to interact and cooperate with people in a shared and constrained environment.

It is a long tradition in robot control and motion planning to focus on the robot's actual safety, i.e., the ability to generate safe paths that avoid collisions with obstacles. However, this is insufficient for robots operating in human congested areas. Studies of human perception have shown that there is a sharp distinction between human perceived safety and the actual safety. This paper presents a path planning framework that takes into account the human’s safety perception in the presence of a flying robot. Human's safety perception is predicted based on data collected from physiological experiments in a virtual reality (VR) environment. In the VR experiment, the participants experience a robot flying in the proximity, and the physiological signals and the position coordinates of the flying robot are recorded simultaneously. The target variable is the physiological arousal signal, and the feature variables are the position and the velocity coordinates of the flying robot.

There are a number of unknown factors present in the data collected from these experiments. Naively assuming the unknown factor to be an independent identically distributed (i.i.d.) Gaussian noise model would not be suitable for the data where only partial observation of the state of the system is provided. However, the i.i.d. Gaussian noise assumption is popularly used for regression tasks, since maximizing the likelihood for estimation task conveniently reduces to the mean squared error (MSE) minimization problem. In~\cite{weninger2014line}, a recurrent neural network (RNN) is applied to predict music mood (valence), where the coefficient of determination\footnote{The coefficient of determination $R^2$ is a performance metric used in regression task. $R^2=1 - \frac{\sum_i(y_i - \hat{y}_i)^2}{\sum_i (y_i - \bar{y})^2}$, where $\bar{y}$ denotes empirical mean and $\hat{y}_i$ denotes the prediction of the target variable given a feature variable $x_i$.}  with a recurrent neural network (RNN) is at most 50\%~\cite{weninger2014line}. The undesired \textit{goodness of fit} despite the highly complex models used in the previous papers suggests that the other factors not contained in the data may influence the outcome of the human test. To overcome the issue, we propose to use a Hidden Markov Model~\cite{ghahramani2001introduction} approach, which divides the data samples into two clusters: (i) relevant samples where the target variable (physiological arousal) can be predicted as a function of the feature (robot's position and velocity); (ii) irrelevant samples where it is better predicted by a random source than a function of the feature. The prediction model estimated with the help of the relevant samples is incorporated in the optimal path planning framework to take into account the human's safety perception. In the optimal path planning, the flight path is parameterized using  Bernstein polynomials, which ensures that the path remains inside the convex hull of its control points~\cite{choe2015trajectory}. This feature helps to have collision avoidance guarantees at low computational cost.

\subsection{BACKGROUND}
Human's perception of a flying robot dependent upon its spatial and temporal variables (e.g. distance and speed) has been studied using virtual reality testbed, and as well using a real flying robot. A comfortable  distance for a flying robot to approach a human was studied in~\cite{duncan2013comfortable}, where the authors tested the effects of the size of the robot on the comfort levels of the human subjects using behavioral, physiological, and survey measures. Distancing with the robot and interaction preference between two differently behaving robots were investigated for speed and repeating behavior (cyclicity) using VR experiments,~\cite{duncan2017effects}.

On the other hand, various design approaches to the \emph{human-aerial-robot} system have been explored again for the purpose of ensuring comfort for humans. Laban effort\footnote{A method to interpret human motion used in choreography\cite{laban1971mastery}} is employed to design affective locomotion for a flying robot in~\cite{sharma2013communicating}, and the effect on arousal and valence due to the design parameter of the locomotion is tested. Emotional encoding in a flight path of a robot was investigated in~\cite{cauchard2016emotion}, where the encoding is derived from characterizing stereotypes of personality and motion parameters using \textit{interaction vocabulary}. In \cite{szafir2015communicating}, the authors propose a flight path design approach, which improves the ease of human's perception of the robot's motion, and the proposed design is tested using survey measures. A signaling device that resembles the turn signals of a car is proposed to communicate the robot's intent to humans,~\cite{szafir2014communication}. Using gestures to communicate the user's intent to the robot is investigated in~\cite{cauchard2015drone}.

In the papers cited above, the focus is on either discovering a general model in \emph{human-aerial-robot} interaction based on the empirical data or devising a heuristic method to improve the acceptability of the robot for humans. However, the sparse and qualitative model stated in the form of null hypothesis testing  is not straightforward to apply for engineers, who intend to use it with an optimal design technique. In the proposed framework, we estimate the uncertain parameters of a human's physiological signal model, then the estimated parametric model is considered as a cost to minimize in the optimal path planning task.

The remainder of the paper is organized as follows. In Section II, the VR experiment set-up is described, and the preliminary finding that relates the physiological arousal to the perceived safety is introduced. In Section III, we propose a model to address the influence of unknown factors and validate it against the VR experimental data.  In Section IV, the optimal path planning that takes into account  the human arousal model is presented. Section V summarizes and discusses future directions.

\section{VR EXPERIMENT AND DATASET GENERATION}
\begin{figure}[thpb]
\centering
 \includegraphics[width=0.4\textwidth]{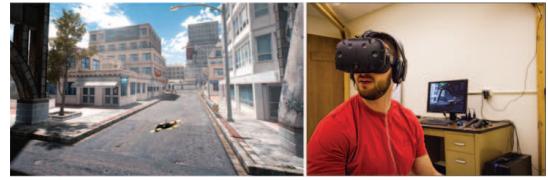}
 \caption{Flying robot observed in the VR environment (an illustration video at \url{https://youtu.be/XnaXzdHlxUA}). }
\medskip
\label{fig:VR_robot}
\end{figure}
Virtual Reality offers a safe, low-cost, and time efficient method to collect data~\cite{duncan2017effects}. For example, the precise coordinates of the human and robot can easily be recorded in real-time, which is useful for studying spatial-temporal variables in human behavior. To this end, we have developed a VR test environment to explore human-aerial-robot interactions in a variety of experimental scenarios~\cite{widdowson2017VR, MarinhoVR2016}. Concurrent psychophysiological reactions of participants are recorded in terms of head motion kinematics and electrodermal activity (EDA), and time-aligned with attributes of the robot's flight path, e.g. velocity, altitude, and audio profile.
\begin{figure}
    \centering
    \begin{subfigure}[b]{0.3\textwidth}
        \includegraphics[width=\textwidth]{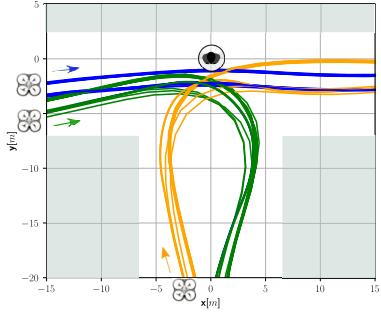}
        \caption{Flight paths.}
        \label{fig:path}
    \end{subfigure}
    ~ 
    \begin{subfigure}[b]{0.35\textwidth}
        \includegraphics[width=\textwidth]{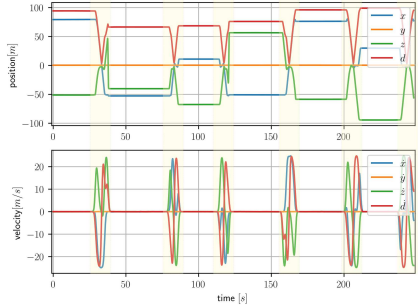}
        \caption{Human-aerial robot interaction events. $x$, $y$, $z$ denote position coordinates and $d$ denotes the distance between the robot and the human. Also, $\dot{x}$, $\dot{y}$, $\dot{z}$ denote velocity coordinates and $\dot{d}$ denotes the rate of change of the distance.}
        \label{fig:robot_events}
    \end{subfigure}
    \caption{Test events and flight paths.}
\end{figure}
During the experiment, participants were introduced to the virtual environment (VE) and told that they would experience a simulation of an urban scene lasting approximately ten minutes. Participants were seated at the junction of a three-way intersection with unoccluded paths in the forward, left, and right direction. Three arbitrary trajectories conforming to the shape of the intersection were chosen and reversed, for a total of six unique trajectories (1.6 m altitude) (see Figure~\ref{fig:path}). The simulation started with a 90 seconds baseline period – without any flying robot – allowing time for the EDA signal to plateau. The first robot then appeared and completed its trajectory. After a pause of 30 to 40 seconds, the next robot appeared, and the process repeated itself for the duration of the entire experiment.  Figure~\ref{fig:robot_events} shows the position and velocity profiles of all these flying robots.  We collected the data from 56 participants (20 males / 36 females) recruited from our university.

\begin{figure}[thpb]
\centering
 \includegraphics[trim={1cm 0.3cm 1cm 1cm},clip,width=0.4\textwidth]{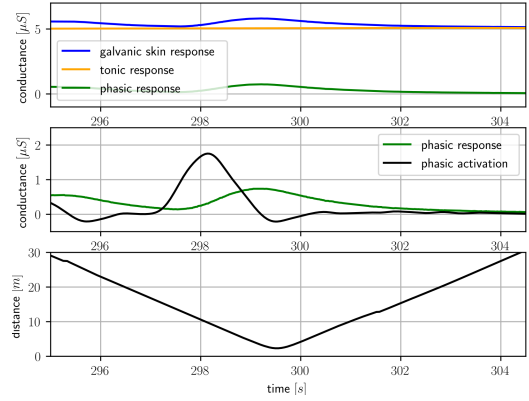}
 \caption{EDA analysis result (phasic/tonic decomposition and deconvolution to determine phasic activation).}
\medskip
\label{fig:EDA}
\end{figure}

The skin conductance signal is preprocessed by EDA analysis package, Ledalab, to generate the phasic activation signal~\cite{benedek2010continuous}. The EDA toolbox decomposes the skin conductance signal into phasic and tonic signal as shown in Figure~\ref{fig:EDA}. The phasic signal is then deconvolved to determine phasic activation; phasic activation represents an instantaneous arousal response.

To our knowledge, there has been no standard index of perceived safety in the literature.  Although physiological measurements of arousal (e.g., EDA) alone are not necessarily equivalent to people's perceived safety, several pieces of evidence suggest that the EDA measure of physiological arousal in our study is closely related to people's anxiety induced by the approaching drone.  For example, in a follow-up experiment examining the effects of path height on people's EDA responses, we found that the EDA phasic response was significantly stronger for a drone approaching at eye-height, where a potential collision was possible, than when the drone was flying at a height beyond the observer’s head, where there is no danger of collision, even though all other characteristics of the drone movement were the same.  These results suggest that such arousal was most likely due to people's anxiety in response to approaching danger rather than general excitement caused by watching flying robots.  Moreover, analysis of the simultaneous head motion showed that as the EDA signal increased with the approaching drone, people made characteristic collision-avoidance movements by jerking their heads away from the drone.  These findings again suggest that the physiological arousal signals observed in our study are most likely a result of people's anxiety to avoid impeding danger, rather than general excitement.  Thus, in the following sections we consider the EDA signal as an operational approximation of the human’s perceived safety for the optimal path generation algorithm.

\section{THE PROPOSED MODEL}
\begin{figure}[thpb]
\centering
 \includegraphics[trim={0.5cm 0.15cm 1cm 1cm},clip,width=0.3\textwidth]{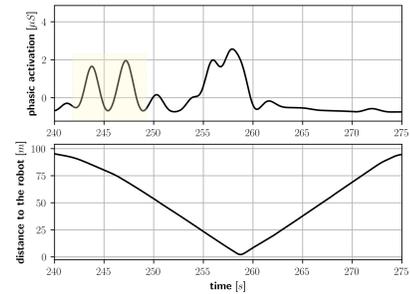}
 \caption{Phasic activation signal induced by the flying robot. The shaded box indicates the response, where the robot is in far distance (greater than 60 [m]).}
 \label{fig:arousal}
\end{figure}

We aim to develop a data-driven model that predicts the phasic activation (arousal), given the robot's position and velocity. Let $y_n\in \mathbb{R}$ denote the phasic activation, where $n$ is the time index. The input (feature) variable, denoted by $x_n \in \mathbb{R}^8$, is the vector that contains the distance to the robot, the rate of change of the distance, the Cartesian position coordinates, and the velocity coordinates.
 Despite the high-fidelity test environment, it is impossible to measure every stimulus on the subject, i.e. there are unknown factors in the data. As an example, one of our collected datasets, shown in  Figure~\ref{fig:arousal}, illustrates the unknown factors' presence in the data. One can notice an increase in the phasic activation in the shaded area, although the flying robot is far away and virtually invisible to the subject.

To account for the unknown factors in the data, we hypothesize that the unexpected spike of the phasic activation in Figure~\ref{fig:arousal} is due to the change of the participant's focus of attention, i.e. the participant is distracted by some other stimulus. Inspired by the work in~\cite{mozer2007sequential}, we model the sequential dependence of the (\emph{hidden}) human's focus of attention using a Hidden Markov Model (HMM). The HMM has two states represented by the latent variable that models the human's attention state, which we denote by
{\small
\begin{equation*}
z_n :=
\left\{
  \begin{array}{@{}ll@{}}
    1, & \text{if the human is attentive to the robot,} \\
    2, & \text{otherwise}.
  \end{array}\right.
\end{equation*}
}
Then $z_n$ is modeled by a homogeneous Markov chain with the following probability transition equation:
{\small
\begin{equation}\label{eq:markov_chain}
\pi_{n+1} = \pi_{n}\mathbf{A}.
\end{equation}
}
 The vector $\pi_k:=[p(z_n=1), p(z_n=2)]$ is the stochastic row vector for the distribution over the state $z_n$, and $\mathbf{A} \in \mathbb{R}^{2 \times 2}$ denotes the transition probability matrix of the Markov chain\footnote{We used $p(\cdot)$ for both probability and probability density; its distinction easily follows from the context.}. The initial distributions $\pi_0$ and $\mathbf{A}$ are the parameters of the Markov chain.

The attention state variable $z_n$ assigns one of the two output emission models $f_{\beta}(x_n)+\epsilon$ or an independent random source $\delta$ as follows
{\small
\begin{equation}\label{eq:model}
	y_n = \ind_{\{z_n=1\}}(f_{\beta}(x_n)+\epsilon) + \ind_{\{z_n=2\}} \delta,
\end{equation}
}
where $\ind_A$ denotes the indicator function, and $f_{\beta}:\mathbb{R}^8 \rightarrow \mathbb{R}$ is a function $f_\beta(x):=\beta^\top \phi(x)$, which is linearly parameterized with $\beta$ and basis $\phi(x)$\footnote{3rd order polynomial basis functions were chosen for the $\phi(\cdot)$.}, $\epsilon \sim \mathcal{N}(0,\sigma^2)$, and $\delta$ denotes the random source. As seen in \eqref{eq:model}, $y_n$ depends on $x_n$ when $z_n=1$; however, $y_n=\delta$ when $z_n=2$, i.e. it is modeled as an independent random signal. In~\eqref{eq:model}, it can be seen that one of the two regression functions of the models $y_n=f_\beta(x_n)$ and $y_n = \delta$ is chosen based on the likelihood given the observation $(x_n, y_n)$.

In addition to the hypothetical  binary HMM that models the change of human's attention, we further hypothesize that the unexpected spike of the activation signal follows a multimodal distribution. We employ a mixture of Gaussians to model $\delta$ for the multimodal distribution. The Gaussian mixture model (GMM) allows a multi-modal and skewed distribution in contrast to the Gaussian distribution. The density of the mixture model for $\delta$ is defined by another latent variable $w_n \in \{1, \dots, K\}$ as follows:
\begin{equation}\label{eq:mix_Gaussian}
p(\delta |w_n=k) = \mathcal{N}(\delta | \mu_k,\,\sigma_k^{2}), \quad p(w_n = k) = c_k,
\end{equation}
where $c_k$ are the mixing coefficients such that $\sum_{k=1}^K c_k=1$, $\mathcal{N}(\delta | \mu,\,\sigma)$ denotes a Gaussian density function of $\delta$ with the mean $\mu$ and the variance $\sigma^2$. We assume that $w_n$ is independent and identically distributed. Also, it is assumed that $w_n$ and $z_n$ are independent, and furthermore, $w_n$ and $z_n$ are conditionally independent, given the observation $(x_n, y_n)$.

\subsection{Model Parameter Estimation}
The model defined by \eqref{eq:markov_chain} - \eqref{eq:mix_Gaussian} has a set of parameters denoted by $\theta:=\{\beta, \mu, \sigma, \pi_1, \mathbf{A}, \{c_i, \mu_i, \sigma_i\}^K_{i=1}\}$.  Given the dataset $\mathbf{x}:=\{x_1, \dots, x_N\}$ and $\mathbf{y}:=\{y_1, \dots, y_N\}$, the parameter of the model is estimated by the maximum likelihood estimation (MLE) through the following conditional likelihood equation:
{\small
\begin{equation}
\argmax_{\theta} p(\mathbf{y}|\mathbf{x}, \theta)
= \argmax_{\theta} \sum_{\mathbf{z}} \sum_{\mathbf{w}} p(\mathbf{y}, \mathbf{z}, \mathbf{w} |\mathbf{x}, \theta),
\end{equation}
}
where the summation takes place over all possible sequences $\mathbf{z}:=\{z_1, \dots, z_N\}$ and $\mathbf{w}:=\{w_1, \dots, w_N\}$. The number of terms for the summation is $2^NK^N$, which makes the optimization intractable for large number of samples. Due to this challenge, Expectation-Maximization (EM) algorithm~\cite{dempster1977maximum} is widely used to obtain the MLE for HMM. \\

\noindent
\textbf{EM Algorithm}:
Assume the complete data $(\mathbf{x}, \mathbf{y}, \mathbf{z}, \mathbf{w})$ is available. Using the independence assumption on $w_n$, $z_n$ and the Markov chain property as defined in~\eqref{eq:markov_chain}, the conditional likelihood is calculated as follows:
{\small
\begin{equation}\label{eq:likelihood}
\begin{aligned}
&p(\mathbf{y}, \mathbf{z}, \mathbf{w} | \mathbf{x}, \theta)= \\
& p(z_1|\pi_1) \left[ \prod_{n=2}^N p(z_n|z_{n-1},\mathbf{A}) \right]\prod_{n=1}^{N}p(w_n |\theta) p(y_n|z_n, w_n, x_n, \theta),
\end{aligned}
\end{equation}
}
where $\mathbf{z}:=\{z_1, \dots, z_N\}$ and $\mathbf{w}:=\{w_1, \dots, w_N\}$. The EM algorithm iteratively maximizes the likelihood using the posterior $p(\mathbf{z}, \mathbf{w} | \mathbf{x}, \mathbf{y}, \theta^{old})$ and the likelihood for the supposed complete data $p(\mathbf{y}, \mathbf{z}, \mathbf{w} | \mathbf{x}, \theta)$, as summarized in Algorithm~\ref{alg:1}. The detailed calculation of EM algorithm for the proposed model is in  Appendix.

In contrast to standard regressions that minimize the mean squared error (MSE), the latent variable model (HMM) determines the parameter of the prediction model $f_\beta(x)$ as the weighted least squared error solution with the weight of the posterior $P(z_{n,1} | \mathbf{x}, \mathbf{y}, \theta^{old})$ as follows:
{\small
\begin{equation*}
\beta^{*}:= \argmin_{\beta}\sum_{n=1}^N P(z_{n,1} | \mathbf{x}, \mathbf{y}, \theta^{old}) (y_n - f_\beta (x_n))^2,
\end{equation*}
}
where $P(z_{n,1} | \mathbf{x}, \mathbf{y}, \theta^{old})$ denotes $P(z_n=1 | \mathbf{x}, \mathbf{y}, \theta^{old})$.
The proposed method puts greater weight on the samples, which are more relevant to the input based on the posterior of the attention state.
{\small
\begin{algorithm}
  \caption{EM Algorithm for MLE of $\theta$ }
  \label{alg:1}
\textbf{Initialize} the parameter, $\theta^{old}$ with $\theta^{0}$.
\begin{algorithmic}
\Repeat{}
  \State{\textbf{1.} Determine the \textbf{posterior},  $p(\mathbf{z}, \mathbf{w} | \mathbf{x}, \mathbf{y}, \theta^{old})$.}
  \State{\textbf{2.} Calculate $Q(\theta, \theta^{old})$:
        \begin{equation*}
            Q(\theta, \theta^{old}):= \sum_{\mathbf{z}, \mathbf{w}}p(\mathbf{z},\mathbf{w}|\mathbf{x}, \mathbf{y}, \theta^{old}) \log p(\mathbf{y}, \mathbf{z}, \mathbf{w} | \mathbf{x}, \theta),
        \end{equation*}
        which is the \textbf{expectation} of $\log p(\mathbf{y}, \mathbf{z}, \mathbf{w} | \mathbf{x}, \theta)$ with respect to the posterior.}
   \State{\textbf{3.} Find the \textbf{maximizer}, $\theta^*$
   \begin{equation*}
        \theta^*  = \argmax_{\theta} Q(\theta, \theta^{old}),
   \end{equation*}}
   \State{\textbf{4.} Update $\theta^{old}$ with $\theta^*$.}
 \Until{convergence.}
\end{algorithmic}
\end{algorithm}
}
\subsection{RESULT}
We choose the following initial parameter $\theta^0$:
{\small
\begin{enumerate}
\item $\beta^{0}:= \argmin_{\beta}\sum_{n=1}^N (y_n - f_\beta (x_n))^2$, and $\sigma^{0}= 0.5$,
\item $\mathbf{A}^{0}:=\begin{bmatrix}1/2 & 1/2\\ 1/2 & 1/2 \end{bmatrix}$, and $\pi_1^{0} := [1/2,1/2]$,
\item $c_k^{0} = 1/K$, $\sigma_k^{0} =1$, and $\mu_k$ are randomly chosen from the interval,$[-1, 1]$,
\end{enumerate}
}
where $K$ is the number of the Gaussian basis of the GMM. For the linear function $f_\beta(x):=\beta^\top \phi(x)$, we choose the basis functions $\phi(x)$ as polynomials with degree 3.

\begin{figure}
 \centering
 \includegraphics[width=0.3\textwidth]{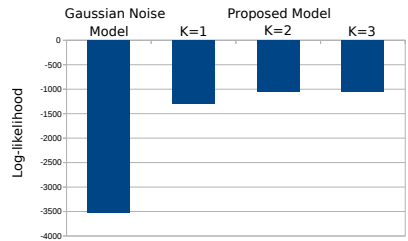}
 \caption{Log-Likelihood with Test Data. $K$ denotes the number of Gaussian basis in the equation \eqref{eq:mix_Gaussian}.}
\label{fig:Log-Likelihood}
\end{figure}

The i.i.d. Gaussian noise model (MSE minimization approach) is contained in the proposed model structure as a special case, $p(z_n=1)=1$, i.e. the arousal is \emph{always} explained by $f_\beta(x)$ than the random source $\delta$. We would like to know whether the proposed model is better than the MSE minimization approach. Likelihood ratio test is typically used to decide if the additional complexity in the modeling is desired compared to a simple model (e.g. i.i.d. Gaussian noise model),~\cite{wasserman2013all}. Since HMM is non-identifiable in general~\cite{yamazaki2003stochastic}, the likelihood ratio model comparison test with the \textit{training dataset} does not apply. To determine which model is more suitable, we calculate the likelihood with \textit{test dataset} by employing the approach from~\cite{uria2016neural}. We randomly partition the data from 56 subjects into a training set with 38 subjects and a test set with the other 18 subjects. Figure~\ref{fig:Log-Likelihood} shows the log-likelihood with test data set for the models: (i) the Gaussian i.i.d. noise model, (ii) the proposed model with a different  basis $K$ of GMM. We see that there is a significant increase in the likelihood using the proposed model as compared to the Gaussian i.i.d. noise model (or the MSE minimization model).
\begin{rem}
 The improvement of the log-likelihood by using the proposed model suggests that the null hypothesis\footnote{The null hypothesis assumes that the true parameter $\varphi$ is in the set of parameters $\Theta_0$, which corresponds to the Gaussian i.i.d. noise model.} is not true. By rejecting the null hypothesis, we show that the proposed model is more suitable than the Gaussian noise model. Consider the following log-likelihood ratio test (see 11.7.4 in~\cite{wasserman2013all}):
{\small
\begin{equation*}
H_0: \varphi \in \Theta_0 \quad \text{versus} \quad H_1: \varphi \in \Theta_1,
\end{equation*}
}
where $\varphi$ denotes the true parameter,  $\Theta_0$ denotes the set of parameters for the Gaussian noise model, and $\Theta_1$ denotes the set of parameters for the proposed model. Note that the Gaussian noise model is a special case of the proposed model, i.e. $\Theta_0 \subseteq \Theta_1$.
The likelihood ratio statistics $\lambda$ is calculated as
{\small
\begin{equation*}
\lambda = 2\log\left(\frac{\sup_{\theta\in\Theta_1}\mathcal{L}(\theta)}{\sup_{\theta\in\Theta_0}\mathcal{L}(\theta)}\right),
\end{equation*}
}
where $\mathcal{L}(\theta):=p(\mathbf{y}, \mathbf{z}, \mathbf{w} | \mathbf{x}, \theta)$ denotes the likelihood for $\theta$ in~\eqref{eq:likelihood}.
Figure~\ref{fig:Log-Likelihood} shows that  $\lambda > 4000$. The relative degree is $r = 10$, as the proposed model has 10 more parameters than the Gaussian noise model. The likelihood ratio test is: reject $H_0$, when $\lambda > \chi^2_{r,\alpha}$. We reject $H_0: \varphi \in \Theta_0 $ with p-value at $0.01$.
\end{rem}

\begin{figure}
    \centering
    \begin{subfigure}[b]{0.4\textwidth}
        \includegraphics[width=\textwidth]{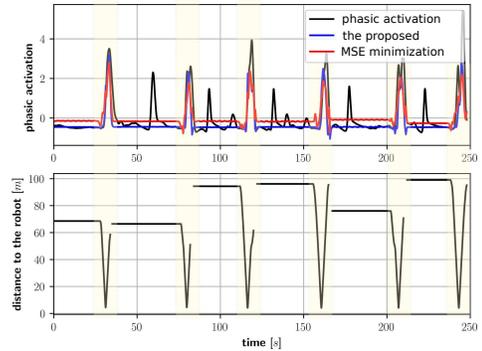}
        \caption{On entire experiment data.}
        \label{fig:Prediction}
    \end{subfigure}
    ~ 
    \begin{subfigure}[b]{0.4\textwidth}
        \includegraphics[width=\textwidth]{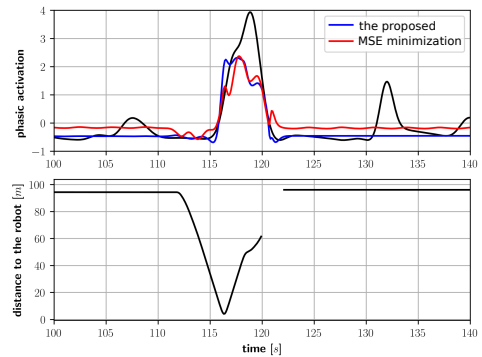}
        \caption{Closer look on an event.}
        \label{fig:Closer_Look}
    \end{subfigure}
    \caption{Prediction, $\hat{\mathbf{y}} = f_\beta(\mathbf{x})$, where the phasic activation signals, are normalized for each subject.}
\end{figure}

We fix the proposed model with $K=2$, since a greater number of basis does not result in significant improvement in likelihood, as shown in Figure~\ref{fig:Log-Likelihood}. The function $f_\beta$ with the fixed model is used to predict the phasic activation (arousal) as shown in Figure~\ref{fig:Prediction}. Figure~\ref{fig:Closer_Look} shows that the MSE minimization approach has signs of over-fitting (oscillation and spiky shape). In the following section, optimal path planning with the prediction model is presented.

\section{OPTIMAL PATH PLANNING}
\begin{figure}[thpb]
\centering
 \includegraphics[trim={0cm 0cm 1cm 1cm},clip,width=0.25\textwidth]{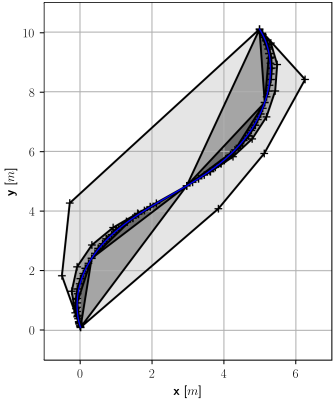}
 \caption{Convex hulls (gray) containing the flight path (blue). Splitting the Bernstein polynomial curve with De Casteljau's algorithm~\cite{farin2000essentials}, we can determine the convex hulls that tightly contain the curve.}
 \label{fig:convexhull}
\end{figure}

The ability to generate a safe path while considering human's safety perception (using the prediction function $f_\beta(\cdot)$) is necessary for our optimal path planning task. To meet the requirement, we employ the trajectory generation method from~\cite{choe2015trajectory, cichella2018optimal}. To ensure spatial separation of the robot's path from obstacles, we use Bernstein polynomials to discretize trajectories.  Bernstein polynomials are useful for checking collision avoidance, as the convex hull of the vertices determined from the coefficients of the Bernstein polynomial contains the  flight path, Figure~\ref{fig:convexhull}. The convex hull is used to check for collision between the vehicle and the obstacles, as demonstrated in~\cite{choe2015trajectory}.

Polynomial interpolation of state trajectories has been used to determine  numerical approximate solutions to optimal path planning. For  example, an interpolation polynomial with the Legendre-Gauss-Lobatto (LGL) time nodes has been used to solve an optimal control problem in~\cite{elnagar1995pseudospectral, ross2012review}. In~\cite{cichella2018optimal} it is shown that using  Bernstein polynomials the  optimal control solution can be approximated sufficiently closely as the number of time-nodes increases, while at the same time ensuring  spatial separation from obstacles.
In our path planning framework, we use the LGL quadrature~\cite{elnagar1995pseudospectral} to approximately calculate the cost to minimize, and we use Bernstein polynomials to check for collision avoidance~\cite{choe2015trajectory}.


\subsection{Finite Dimensional Optimization}
Both Bernstein polynomial curves and interpolation polynomial curves with LGL nodes are two equivalent parametrizations for a polynomial trajectory. Consider a 2D trajectory time function $\vect{p}(t): = [x(t), y(t)]^\top$ as an $n^{th}$ order polynomial in $x(t)$ and $y(t)$. We present a brief overview of both equivalent representations.
\begin{enumerate}
	\item A degree $n$ Bernstein polynomial is given by:
	{\small
	\begin{equation*}  
		\vect{p}(t)
		=
		\sum^{n}_{k=0}
		\bar{\vect{p}}_k
		b^n_k(\zeta(t)), \quad
		\zeta: [0, t_f] \rightarrow [0,1],
		\quad
		\zeta(t):= \frac{t}{t_f},
	\end{equation*}
	}where  $b^n_k(\zeta):= \binom{n}{k}(1-\zeta)^{n-k} \zeta^k, \quad \zeta \in [0,1]$, represents the polynomial basis, and the coefficients $\bar{\vect{p}}_k$ are called control points\footnote{Note that the polynomial is a vector equation, so  the coefficient $\bar{\vect{p}}_k$ is also a vector.} of the Bernstein polynomial.
	
	\item  The interpolation curve  is represented as:
	{\small
	\begin{equation*}
		\vect{p}(t)
		=
		\sum^{n}_{k=0}
		\vect{p}_k
		\ell_k(\eta(t)),
		\quad
		\eta: [0, t_f] \rightarrow [-1,1],
		\quad
		\eta(t):= \frac{2t}{t_f}-1,
	\end{equation*}
	}where $\vect{p}_k$ are interpolation points at time nodes $t_k$, and $\ell_k(\eta(t)):= \prod_{0\leq i \leq n, i \neq k } \left(\frac{\eta(t) - \eta(t_i)}{\eta(t_k) - \eta(t_i)}\right)$ are the Lagrange polynomial basis.
\end{enumerate}
The $n^{th}$ order polynomial trajectories can be parameterized by either $n+1$ control points $\bar{\vect{p}}_k$, or $n+1$ interpolation points $\vect{p}_k$, and the transformation between the control points and the interpolation points can be done using matrix multiplication.

The optimal path planning is formulated as the following finite dimensional optimization:
\begin{equation*} \label{optimal_control}
 \begin{aligned}
                    & \argmin_{\bar{\vect{p}}_0, \dots,  \bar{\vect{p}}_n, t_f} J(\bar{\vect{p}}_0, \dots,  \bar{\vect{p}}_n, t_f),\\
  \text{subject to }& \text{collision avoidance constraint}, \\
                    & \text{velocity and acceleration constraint},\\
 \end{aligned}
\end{equation*}
where $J(\bar{\vect{p}}_0, \dots,  \bar{\vect{p}}_n,  t_f)$ is the LGL quadrature of $\int_{0}^{t_f}  L(\vect{p}(t), \dot{\vect{p}}(t))dt$ calculated by the method in~\cite{elnagar1995pseudospectral}, and $L(\vect{p}(t), \dot{\vect{p}}(t))$ is the running cost to be minimized in time average. Constraint equations for collision avoidance and velocity acceleration can be written as  functions of control points $\bar{\vect{p}}_0, \dots,  \bar{\vect{p}}_n$ by following the methodology in~\cite{choe2015trajectory}.

\subsection{Optimal Path Planning in the Presence of Humans}
Define $\vect{x}(t)$ in the same way as we defined $x_n$ in~\eqref{eq:model}, where $\vect{x}(t) \in \mathbb{R}^8$ contains the distance to the robot, the rate of change of the distance, the position coordinates and the velocity coordinates at time $t$. Notice that with the polynomial path $\vect{p}(t)$ and $\dot{\vect{p}}(t)$ one can directly construct $\vect{x}(t)$. For this reason, to simplify the notation we can use $\vect{x}(t)$ and $(\vect{p}(t), \dot{\vect{p}}(t))$ interchangeability as arguments of the functions $f_\beta(\cdot),\: J(\cdot)$ and $L(\cdot)$.

In the optimal path planning, we only consider values of $f_\beta$ larger than a threshold $b_a$, where $b_a~\ge~0$ is essentially a tuning parameter. Intuitively, we ignore arousal levels below the threshold. To make the optimization problem tractable, instead of adding a strict constraint to the minimization problem, the constraint is incorporated in the running cost as a penalty function~\cite{zangwill1967non}:
{\small
\begin{equation} \label{eq:running_cost}
        L(\vect{p}(t), \dot{\vect{p}}(t)):= 1 + \gamma \max(0, f_\beta(\vect{x}(t)) - b_a)^2,
\end{equation}
}
where $\gamma$ is the penalty coefficient. The corresponding cost function $J(\bar{\vect{p}}_0, \dots,  \bar{\vect{p}}_n,  t_f)$ becomes
{\small
\begin{equation} \label{eq:the_cost}
    J(\bar{\vect{p}}_0, \dots,  \bar{\vect{p}}_n,  t_f) = t_f + \gamma \int_{0}^{tf} \max(0, f_\beta(\vect{x}(t)) - b_a)^2 dt.
\end{equation}
}

\begin{figure}[thpb]
\centering
 \includegraphics[width=0.35\textwidth]{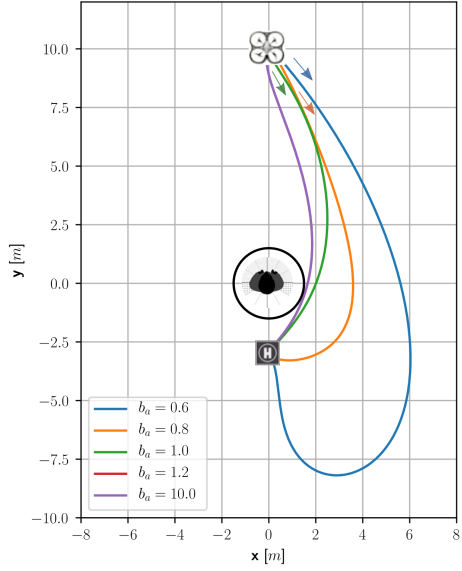}
 \caption{Flight paths generated with the proposed model.}
 \label{fig:Optimal flight paths_proposed}
\end{figure}

\begin{figure}[thpb]
\centering
 \includegraphics[width=0.35\textwidth]{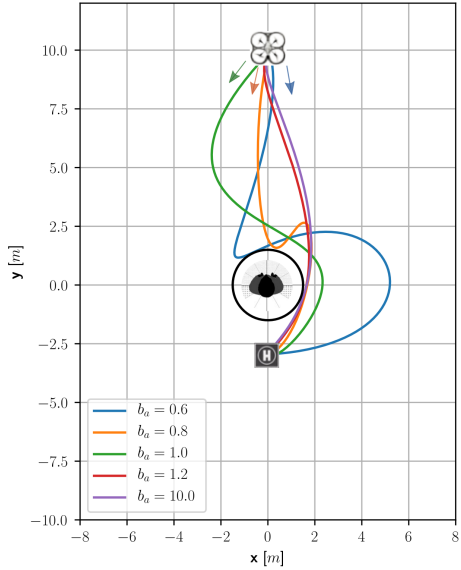}
 \caption{Flight paths generated with the MSE minimization model.}
 \label{fig:Optimal flight paths_deterministic}
\end{figure}

The two arousal prediction functions are used in the optimal flight trajectory generation, as shown in Figure~\ref{fig:Optimal flight paths_proposed} and Figure~\ref{fig:Optimal flight paths_deterministic}. The smaller value of $b_a$  results in a path that is more safety conscious, as intended by the running cost function in~\eqref{eq:running_cost}. Flight paths generated with the proposed model show the desirable behavior, as shown in Figure~\ref{fig:Prediction} (decreasing $b_a$ results in greater distance from the human).
However, the paths with the MSE minimization model have unconvincing shapes. This undesirable behaviour of the MSE minimization model is due to over-fitting, as we have seen in Figure~\ref{fig:Closer_Look}. It shows that the arousal prediction model which only minimizes MSE does not generalize in the optimal path generation task.

\section{CONCLUSION}
We present a path planning framework that takes into account  human's safety perception. Psychophysiological reactions for different paths of flying robots in VR were collected to estimate the arousal prediction model. To consider the unknown factor in the data, we proposed a hidden Markov model approach. Compared to the mean squared error (MSE) minimization approach (due to i.i.d. Gaussian noise model), the proposed model has improved the likelihood significantly. When the arousal prediction functions are implemented in the optimal path planning, the flight paths with the proposed model show desirable behaviors, in contrast to the unconvincing flight paths with the MSE minimization approach.
\section*{APPENDIX}
The EM algorithm for the proposed model is calculated by the following three subsequent steps:

\subsubsection{\textbf{Determine the Posterior} $p(\mathbf{z}, \mathbf{w} | \mathbf{x}, \mathbf{y}, \theta^{old})$}
Using the conditional independence assumption of the latent variables, the posterior is factorized as
\begin{equation} \label{eq:factored_post}
p(\mathbf{z}, \mathbf{w} | \mathbf{x}, \mathbf{y}, \theta^{old})=p(\mathbf{z} | \mathbf{x}, \mathbf{y}, \theta^{old})p(\mathbf{w} | \mathbf{x}, \mathbf{y}, \theta^{old}).
\end{equation}
Now, we can calculate $p(\mathbf{z}| \mathbf{x}, \mathbf{y}, \theta^{old})$ and $p(\mathbf{w}| \mathbf{x}, \mathbf{y}, \theta^{old})$ separately. \\

\noindent
\textbf{Forward-Backward Algorithm}~\cite{bishop2006pattern}:\\
Define $a(z_{n,i})$ and $b(z_{n,i})$ as follows:
\begin{equation*}
\begin{aligned}
a(z_{n,i}) &:= p(y_1, \dots, y_n, z_{n,i}), \\
b(z_{n,i}) &:= p(y_{n+1}, \dots, y_N | z_{n,i}, \mathbf{x}),
\end{aligned}
\end{equation*}
where $z_{n,i}$ denotes the event $\{z_n = i\}$. To calculate the posterior, the following recursive equations are used:
{\small
\begin{equation*}\label{eq:forward-backward}
\begin{aligned}
a(z_{n,i}) &= p(y_n|z_{n,i}, \mathbf{x}, \theta^{old}) \sum_{k=1}^{2} a(z_{n-1,k})p(z_n|z_{n-1,k}, \mathbf{A}^{old}),\\
b(z_{n,i}) &= \sum_{k=1}^{2} b(z_{n+1,k})p(y_{n+1}|z_{n+1,k}, \mathbf{x}, \theta^{old})p(z_{n+1,k}|z_{n,i}, \mathbf{A}^{old}),
\end{aligned}
\end{equation*}
}where $p(y_{n}|z_{n,i}, \mathbf{x}, \theta^{old})$ is calculated using~\eqref{eq:model} as
\begin{equation*}
\begin{aligned}
&p(y_{n}|z_{n,i}, \mathbf{x}, \theta)
 \\
&=
\left\{
  \begin{array}{@{}ll@{}}
   \mathcal{N}(y_n - f_\beta (x_n) | 0,\,\sigma^{2}), & \text{if}\ i=1 \\
    \sum_{k=1}^K \phi_k \mathcal{N}(y_n  | \mu_k,\,\sigma_k^{2}), & \text{if}\ i=2.
  \end{array}\right.
\end{aligned}
\end{equation*}

The boundary values $a(z_1)$ and $b(z_N)$ are determined as $a(z_1)=p(z_1|\pi_1)p(y_1|\mathbf{x},\theta^{old})$ and $b(z_N) = 1$. After calculating $a(z_n)$ recursively and $b(z_n)$, the posterior is determined as follows:
{\small
\begin{equation}\label{eq:HMM_post}
p(z_n|\mathbf{x}, \mathbf{y}, \theta^{old} ) = \frac{a(z_n)b(z_n)}{p( \mathbf{y} |\mathbf{x}, \theta^{old})},
\end{equation}
\begin{equation}
p(z_{n-1, j}, z_{n, k} | \mathbf{x}, \mathbf{y}, \theta^{old}) = \frac{a(z_{n-1,j})p(y_n|z_{n,k},x_n,\theta^{old})A_{jk}b(z_{n,k})}{p( \mathbf{y} |\mathbf{x}, \theta^{old} ) },
\end{equation}
}where the likelihood is calculated as
\begin{equation}\label{eq:overall_likelihood}
 p( \mathbf{y} |\mathbf{x}, \theta^{old}) = \sum_{k=1}^{2} a(z_{N,k}).
\end{equation}\\
\noindent
\textbf{Posterior for the GMM}: Due to the conditional independence assumption of $w_n$ and $z_n$, the posterior $p(w_{n,i}|x_n, y_n, \theta^{old})$ is calculated independently from $p(z_n|\mathbf{x}, \mathbf{y}, \theta^{old} )$, by directly using the result on GMM from~\cite{bishop2006pattern} as follows
\begin{equation}\label{eq:GMM_posterior}
p(w_{n,i}|x_n, y_n, \theta^{old}) = \frac{\phi_i \mathcal{N}(y_n|\mu_i, \sigma_i^2)}{\sum_{k=1}^{K}\phi_k \mathcal{N}(y_n|\mu_k, \sigma_k^2)},
\end{equation}
where $w_{n,i}$ denotes the event $\{w_n=i\}$.

\subsubsection{\textbf{Calculate} $Q(\theta, \theta^{old})$}
Using the poseterior calculated in~\eqref{eq:HMM_post} and~\eqref{eq:GMM_posterior} and the likelihood in~\eqref{eq:likelihood},   $Q(\theta, \theta^{old})$ in Algorithm~\ref{alg:1} is calculated by expanding the log term:
{\small
\begin{equation} \label{eq:Q}
\begin{aligned}
&Q(\theta, \theta^{old})\\
& := \sum_{z, w}p(\mathbf{z},\mathbf{w}|\mathbf{x}, \mathbf{y}, \theta^{old}) \log p(\mathbf{y}, \mathbf{z}, \mathbf{w} | \mathbf{x}, \theta) \\
& = \sum_{i=1}^2 p(z_{1,i}| \mathbf{x}, \mathbf{y}, \theta^{old}) \log \pi_{1,i} \\
+& \sum_{n=2}^{N} \sum_{i=1}^{2} \sum_{j=1}^{2} p(z_{n-1, i}, z_{n, j} | \mathbf{x}, \mathbf{y}, \theta^{old}) \log A_{i,j} \\
+& \sum_{n=1}^{N} \sum_{i=1}^{2} \sum_{k=1}^{K} p(z_{n,i}, w_{n,k}| \mathbf{x}, \mathbf{y}, \theta^{old})  \log p(w_n | \phi_k) p(y_n |z_{n,i}, w_{n,k}, x_n, \theta),
\end{aligned}
\end{equation}
}where $z_{n,i}$ denotes the event $\{z_n = i\}$,  $w_{n,k}$ denotes the event $\{w_n =k\}$,  $A_{i,j}$ is the $(i,j)$ element of the matrix $\mathbf{A}$, and $ p(y_n |z_{n,i}, w_{n,k}, x_n, \theta)$ is calculated using the model equation \eqref{eq:model} as follows
\begin{equation}\label{eq:p_y}
\begin{aligned}
&p(y_n |z_{n,i}, w_{n,k}, x_n, \theta) \\
&=
\left\{
  \begin{array}{@{}ll@{}}
    \mathcal{N}(y_n - f_\beta (x_n) | \mu,\,\sigma^{2}), & \text{if}\ i=1 \\
     \mathcal{N}(y_n  | \mu_k,\,\sigma_k^{2}), & \text{if}\ i=2.
  \end{array}\right.
\end{aligned}
\end{equation}

\subsubsection{ \textbf{Find the maximizer} $\theta^*$}
As seen in~\eqref{eq:Q} and~\eqref{eq:p_y}, each term has a distinct set of parameters. Hence, we can determine the maximizer for each term independently from the other terms. From~\cite{bishop2006pattern} we have the maximizers as follows:
{\small
\begin{equation}\label{eq:pi1_update}
\pi_{1,i}^{*} = \frac{p(z_{1,i}| \mathbf{x}, \mathbf{y}, \theta^{old})}{\sum_{j=1}^{2}p(z_{1,j}| \mathbf{x}, \mathbf{y}, \theta^{old})},
\end{equation}
\begin{equation}\label{eq:A_update}
A_{j,k}^{*} = \frac{\sum_{n=2}^{N}p(z_{n-1, j}, z_{n, k} | \mathbf{x}, \mathbf{y}, \theta^{old})}{\sum_{l=1}^{2}\sum_{n=2}^{N}p(z_{n-1, j}, z_{n, l} | \mathbf{x}, \mathbf{y}, \theta^{old})}.
\end{equation}
}

The maximizer for the last term in~\eqref{eq:Q} is calculated using the model equation \eqref{eq:model}. Let $L$ denote the last term in~\eqref{eq:Q}. Using \eqref{eq:p_y}, $L$ is written as follows:
{\footnotesize
\begin{equation*} \label{eq:L}
\begin{aligned}
&L(\beta, \sigma, \{\phi_i, \mu_i, \sigma_i\}^K_{i=1})\\
&:= \sum_{n=1}^{N} \sum_{i=1}^{2} \sum_{k=1}^{K} p(z_{n,i}, w_{n,k}| \mathbf{x}, \mathbf{y}, \theta^{old})  \log p(w_n | \phi_k) p(y_n |z_{n,i}, w_{n,k}, x_n, \theta) \\
& = \sum_{n=1}^{N} \sum_{k=1}^{K} p(w_{n,k}| \mathbf{x}, \mathbf{y}, \theta^{old}) \log \phi_k\\
& + \sum_{n=1}^{N} p(z_{n,1} | \mathbf{x}, \mathbf{y}, \theta^{old}) \log \left( \frac{1}{\sigma\sqrt{2\pi}}\exp\left(-\frac{(y_n - f_\beta(x_n))^2}{2 \sigma^2}\right)   \right) \\
& + \sum_{n=1}^{N} \sum_{k=1}^{K} p(z_{n,2}, w_{n,k}| \mathbf{x}, \mathbf{y}, \theta^{old}) \log \left( \frac{1}{\sigma_k\sqrt{2\pi}}\exp\left(-\frac{(y_n - \mu_k)^2}{2 \sigma_k^2}\right)   \right).
\end{aligned}
\end{equation*}
}
The variable $\beta^{*}:=\argmax_{\beta}Q(\beta, \theta^{old})$ is calculated as
\begin{equation}\label{eq:beta_star}
\beta^{*}:= \argmin_{\beta}\sum_{n=1}^N P(z_{n,1} | \mathbf{x}, \mathbf{y}, \theta^{old}) (y_n - f_\beta (x_n))^2.
\end{equation}

The $\phi_i^{*}$, $\mu_i^{*}$, $\sigma_i^{*}$, and $\sigma^{new}$ are determined using KKT (Karush-Kuhn-Tucker) condition as follows:
{\small
\begin{equation*}
\phi_i^{new} = \frac{\sum_{n=1}^{N} p(w_{n,i}| \mathbf{x}, \mathbf{y}, \theta^{old})}{\sum_{n=1}^{N} \sum_{k=1}^{K} p(w_{n,k}| \mathbf{x}, \mathbf{y}, \theta^{old})},
\end{equation*}

\begin{equation*}
\mu_i^{new} = \frac{\sum_{n=1}^{N} p(z_{n,2},w_{n,i}| \mathbf{x}, \mathbf{y}, \theta^{old}) \, y_n }{\sum_{n=1}^{N} p(z_{n,2},w_{n,i}| \mathbf{x}, \mathbf{y}, \theta^{old})},
\end{equation*}

\begin{equation*}
\sigma_i^{new} = \frac{\sum_{n=1}^{N} p(z_{n,2},w_{n,i}| \mathbf{x}, \mathbf{y}, \theta^{old}) \, (y_n - \mu_i^{new})^2 }{\sum_{n=1}^{N} p(z_{n,2},w_{n,i}| \mathbf{x}, \mathbf{y}, \theta^{old})},
\end{equation*}

\begin{equation*}
\sigma^{new} = \frac{\sum_{n=1}^{N} p(z_{n,1}| \mathbf{x}, \mathbf{y}, \theta^{old}) \, (y_n - f_{\beta^{new}}(x_n))^2 }{\sum_{n=1}^{N} p(z_{n,1}| \mathbf{x}, \mathbf{y}, \theta^{old})}.
\end{equation*}
}



\bibliographystyle{IEEEtran}
\bibliography{mybibfile}

\end{document}